\documentclass[a4paper,twoside]{article}

\usepackage{epsfig}
\usepackage{subcaption}
\usepackage{calc}
\usepackage{amssymb}
\usepackage{amstext}
\usepackage{amsmath}
\usepackage{amsthm}
\usepackage{multicol}
\usepackage{multirow}
\usepackage{pslatex}
\usepackage{apalike}
\usepackage{algorithm2e}
\usepackage[bottom]{footmisc}

\usepackage{array}   
\newcolumntype{L}{>{$}l<{$}} 
\newcolumntype{R}{>{$}r<{$}} 
\newcolumntype{C}{>{$}c<{$}} 

\usepackage{lipsum}
\usepackage{xcolor}

\usepackage{url}
\usepackage[capitalize]{cleveref}
\crefname{section}{Sec.}{Secs.}
\Crefname{section}{Section}{Sections}
\Crefname{table}{Table}{Tables}
\crefname{table}{Tab.}{Tabs.}

\usepackage{booktabs}

\usepackage{SCITEPRESS}     
\begin{document}

\def\trade{TrADe}
\def\botsort{BotSORT}
\def\bytetrack{ByteTrack}
\def\hybridsort{HybridSORT}
\def\ocsort{OC-SORT}
\def\deepocsort{Deep OC-SORT}
\def\strongsort{StrongSORT}
\def\supertracker{WindowTracker}
\def\idfone{$IDF1$}
\def\hota{$HOTA$}
\def\mota{$MOTA$}
\def\motp{$MOTP$}
\def\clip{CLIP-ReiD}
\def\osnet{OSNet}

\def\reidmodel{OSNet~\cite{Zhou2019}}

\def\dataset{UFPR-Planalto801}

\title{A Multilevel Strategy to Improve People Tracking in a Real-World Scenario}

\author{
\authorname{
Cristiano B. de Oliveira \sup{1,2}\orcidAuthor{0000-0003-2320-2724}, 
Joao C. Neves \sup{3}\orcidAuthor{0000-0003-0139-2213}, 
Rafael O. Ribeiro \sup{4}\orcidAuthor{0000-0002-6381-3469} and David Menotti \sup{1}\orcidAuthor{0000-0003-2430-2030}}
\affiliation{
\sup{1}Department of Informatics, Federal University of Paran\'{a}, Curitiba, Brazil}
\affiliation{
\sup{2}Federal University of Cear\'{a}, Quixad\'{a}, Brazil}
\affiliation{
\sup{3}University of Beira Interior, NOVA-LINCS, Portugal}
\affiliation{
\sup{4}National Institute of Criminalistics, Brazilian Federal Police, Curitiba, Brazil}
\email{cristianobac@ufc.br, jcneves@di.ubi.pt, rafael.ror@pf.gov.br, menotti@inf.ufpr.br}
}


\abstract{
The Pal\'acio do Planalto, office of the President of Brazil, was invaded by protesters on January 8, 2023. Surveillance videos taken from inside the building were subsequently released by the Brazilian Supreme Court for public scrutiny.
We used segments of such footage to create the UFPR-Planalto801 dataset for people tracking and re-identification in a real-world scenario. This dataset consists of more than 500,000 images. 
This paper presents a tracking approach targeting this dataset. 
The method proposed in this paper relies on the use of known state-of-the-art trackers combined in a multilevel hierarchy to correct the ID association over the trajectories.
We evaluated our method using IDF1, MOTA, MOTP and HOTA metrics. 
The results show improvements for every tracker used in the experiments, with IDF1 score increasing by a margin up to $9.5$\%.
}

\onecolumn \maketitle \normalsize \setcounter{footnote}{0} \vfill

\section{\uppercase{Introduction}}
\label{sec:intro}
\noindent

On January 8, 2023, the Palácio do Planalto, main office of the President of Brazil, was invaded by protesters who alleged that there was fraud in the presidential elections.
Nearly three months after these events, the Brazilian Supreme Court released all the footage from that day
to the public access \cite{GSI2023}.
The released footage is composed by surveillance videos\footnote{Available for download in \url{https://drive.presidencia.gov.br/public/615ba7}}
that show the activities in several locations inside the building during the whole day.
We used pieces of  such footage to create the \textit{UFPR-Planalto801 dataset} 
\footnote{\url{https://web.inf.ufpr.br/vri/databases/ufpr-planalto801}}.
This dataset contains images of a real-world surveillance scenario and it is mainly intended for use in the development of security systems, specially regarding people tracking and re-identification.

On constructing this dataset, we conducted experiments by using state-of-the-art tracking approaches on the selected pieces of footage. Each of the used approaches presented flaws on keeping consistent trajectories, mainly due to a high number of miss identified people.
Therefore, alongside with the \dataset\ dataset (\Cref{sec:dataset}), this paper presents \supertracker, a tracking strategy (\Cref{sec:proposal}) that combines pairs of trackers into a multilevel hierarchy to correct the ID association over the trajectories.

For the experiments (\Cref{sec:setup}) we used six different trackers and combined them in a total of twelve pairs. The trackers are:  \bytetrack~\cite{zhang2022bytetrack}, \botsort~\cite{aharon2022bot}, \strongsort~\cite{du2023strongsort}, \ocsort~\cite{Cao_2023_CVPR}, \deepocsort~\cite{maggiolino2023deep} and \hybridsort~\cite{yang2023hybrid}.  
All these trackers are able to work with online and real-time videos, and have achieved good results running on datasets like DanceTrack~\cite{Sun_2022_CVPR} and MOTChallenge~\cite{MOT19_CVPR}\cite{MOTChallenge20}.

Each of the pairs was then compared to the individual trackers in order to assess the effectiveness of this pairing system.
The results (\Cref{sec:results}) show improvements for every tracker used in the experiments, with IDF1~\cite{Ristani2016} score increasing by a margin from $2.1$\% (\bytetrack\ and \botsort) up to $9.5$\% (\deepocsort).

\section{\uppercase{People Tracking}}
\label{sec:fundamentals}
\noindent
The Multiple Object Tracking (MOT) problem entails the simultaneous tracking of multiple objects in a video. This requires a manner for identifying them consistently throughout the frames in order to establish their trajectories (or tracks) along the video. 
Pedestrian tracking in video constitutes a specific instance of MOT wherein the objects of interest are the individuals present in the video.

A common paradigm used in MOT tasks is to split the problem in two main steps, which are detecting the objects each frame and then associate them to previous detected objects.
This association occurs across frames by tagging all detected objects with an unique ID number, in such way that all objects that share the same ID are considered the same tracked object.
Errors from both detection and association steps impact the overall results of the tracking process. This paradigm is known as Tracking by Detection (TBD). 

TBD trackers often employ techniques like  Kalman filters (KF) for modeling motion and to estimate objects positions based on previously computed trajectories. Besides avoiding the tracker to work with a batch of detections regarding an object, such kind of modeling also helps it to correct detection errors.

In addition to motion prediction, trackers also rely on features gathered from the bounding boxes surrounding the detected objects. Such features are usually considered to perform the re-identification of objects for the association step. 
There are several re-identification proposals in the literature, ranging from those based on metrics such as Intersection over Union (IoU) to more complex deep learning models.

\section{\uppercase{Related Work}}
\label{sec:related}
\noindent
\bytetrack\ is a tracking method that  executes a two-step association process in order to use all detected bounding boxes to determine the objects tracklets.
At first, \bytetrack\ only tries to match bounding boxes with high-confidence values while the remaining ones are used in a subsequent step. The algorithm uses IoU as a distance metric for the association between the detected and predicted objects.
When associating the low-confidence bounding boxes, \bytetrack\ tries to identify which ones correspond to object detections and which correspond to the background of the image by computing their similarities with tracklets that have not yet been associated. 

In addition to \bytetrack, this paper works with a number of state-of-the-art methods for pedestrian tracking based on the SORT algorithm~\cite{Bewley2016}. These methods use strategies similar to \bytetrack\ \cite{zhang2022bytetrack}, employing more than one step for associating the bounding boxes.

SORT (Simple Online and Real-time Tracking) is a tracking algorithm designed for real-time applications. SORT is known for its simplicity, speed, and real-time tracking capabilities in videos. It implements an estimation model to keep the tracking state via KF. The assignment of the detections to the tracks relies on a cost matrix of IoU distances between detected and predicted bounding boxes.
SORT may struggle in complex scenarios with heavy occlusion, no-linear movement or frequent appearance changes. These situations may require modifications on SORT, which has inspired other tracking approaches.

\botsort~\cite{aharon2022bot} derives from SORT by introducing modifications to the state vector in KF.
Such modifications aims to improve the fitting of bounding boxes by including values for their height and width.
Another difference regarding SORT is the use of a camera motion compensation module based on optical flow. It also relies on appearance Re-ID descriptors from a ResNet50 backbone network along with the IoU distances to perform associations.

\strongsort~\cite{du2023strongsort} is mainly derived from DeepSORT~\cite{DeepSORT2017}, which was 
one of the earliest SORT-like method that applied deep learning to tracking.
A key difference between \strongsort\ and Deep SORT is the detector they use. While Deep SORT uses a CNN with two convolution layers, \strongsort\ uses the YOLOX~\cite{ge2021yolox} detection model.
In terms of motion prediction, \strongsort\ also differs from Deep SORT by using a Kalman NSA filter~\cite{GIAOTracker2021}. Two other features implemented in \strongsort\ are the Camera Motion Compensation and Exponential Moving Average modules. The former uses e technique to improve correlation coefficient maximization, and the latter tries to enhance long-term association.

\ocsort~\cite{Cao_2023_CVPR} is a method designed to enhance tracking robustness under occlusion and non-linear object motion (when objects exhibit variable velocities within a given time interval). It includes the strategy of generating updated iterations of the KF based on virtual trajectories created by a module referred as "Observation-Centric Re-Update" (ORU), which relies on historical observations. ORU is meant to reduce occlusion error accumulation over time, particularly when objects are not under active tracking. Objects that were not tracked for an interval can be revisited, allowing the reactivation of previously inactive tracklets.
\ocsort\ tackles non-linear motion by inserting a term denoted as "Observation-Centric Momentum" (OCM) within the association cost matrix, computed considering the direction of object motion during the association process. 

\deepocsort~\cite{maggiolino2023deep} modifies \ocsort\ for enhancing the precision and robustness of associations. Its tracking model is tuned with dynamic information about the visual appearance of objects. 
\deepocsort\ identifies situations such as occlusion or blur based on detections confidence. Detections with low values are rejected during the similarity cost computation.
Thus, the process is adapted to amplify the significance of appearance features considering instances of high-quality detections.

The \hybridsort\ \cite{yang2023hybrid} central idea is to modify strategies used in other trackers to include information typically considered less significant. \hybridsort\ considers that the object's height is related to camera depth, while the confidence value can indicate the occurrence of occlusions, a premise similar to what is incorporated in \deepocsort.
Thus, it modifies the OCM module of \ocsort\ to include height and confidence, proposing two new modules: \textit{Tracklet Confidence Modeling} and \textit{Height Modulated IoU}.
\hybridsort\ uses the same two-step association strategy as \bytetrack\ but includes a third step to reactivate tracklets, similar to \ocsort. Like \strongsort, \hybridsort\ includes an EMA module, using cosine distance to calculate the similarity between the predicted and detected features.

\section{\uppercase{Proposed Dataset}}
\label{sec:dataset}

\begin{figure}[ht]
  \centering
   \includegraphics[width=.9\linewidth]{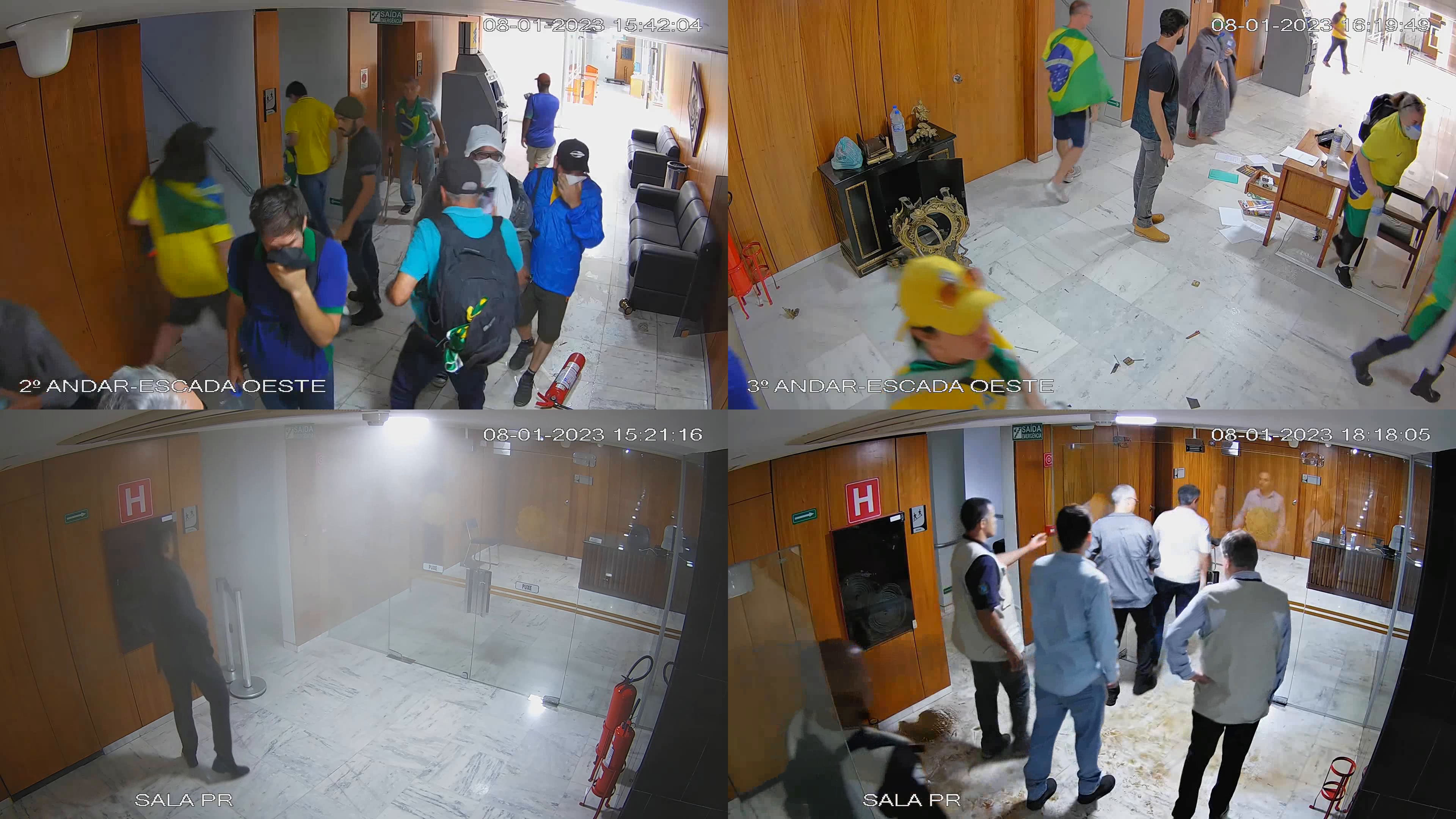}

   \caption{Examples of scenes captured on footage. }
   \label{fig:examples}
\end{figure}

\noindent
The \dataset\ dataset is composed by videos taken from several locations inside the Palácio do Planalto, during protesters invasion on a Sunday, January 8, 2023.
These videos are part of the footage released to the public by Brazilian Supreme Court.
Despite the releasing of 1557 videos recorded that day, videos taken before the invasion occurs (around 3 pm) typically show still images of empty rooms. This dataset is intended for tasks related to people tracking and therefore we selected and clipped parts of videos in order to gathering interesting pieces of footage on people activities after the invasion time. \Cref{fig:examples} shows examples of scenes from footage, showing images from 3 different rooms, besides situations with the presence of smoke and glass reflections.

The \dataset\ dataset contains $14$ videos of several lengths, with a resolution of 1920x1080 pixels and encoded to 24 FPS. There are approximately 6 hours of video, with a total of $518050$ frames and $510471$ annotated detections. 
\Cref{tab:mot_datasets} shows a comparison between \dataset\ and others datasets commonly used in MOT.

\begin{figure}[ht]
  \centering
   \includegraphics[width=.9\linewidth]{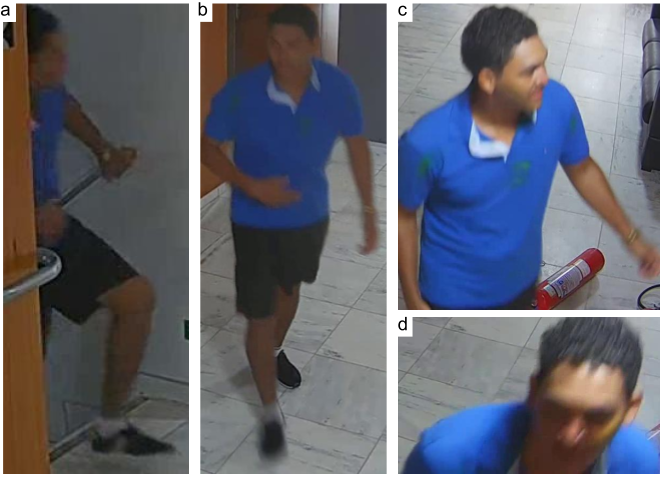}
   \caption{Framing of people in the proposed dataset: (a) occluded; (b) full body; (c) upper body; and (d) head only.}
   \label{fig:bodyparts}
\end{figure}

In order to preserve the complexity of a real-world scenario, we kept detections of people within irregular framing (as in \Cref{fig:bodyparts}).
This increases the challenge on working with \dataset\ and also differentiates it from others datasets. For a matter of comparison, people in DanceTrack~\cite{Sun_2022_CVPR} are mainly framed in full body (with few exceptions).

In all videos of the \dataset\ dataset there are frames with no detections. Such frames were not removed in order to preserve the videos integrity, since during a normal ongoing footage there will not be any video editing.
Besides that, in many situations a person who left the scene reappears after a few frames in a very unlikely location (\textit{e.g.}, a different entrance). This can be challenging because it often results in errors in tracking state, leading the tracker to misidentify this person and to start a new tracklet, what increases the number of track fragments.

\renewcommand{\arraystretch}{1.1}

\begin{table}[ht]
\fontsize{9pt}{9pt}\selectfont
\centering
\addtolength{\tabcolsep}{-3.5pt}
\caption{Comparison between \dataset\ and other commonly used MOT datasets (approximated values).}

\begin{tabular}{|p{2.75cm}|C|C|C|C|}
\hline
 Dataset & Frames & Minutes & Boxes&Tracks\\

\hline
 DanceTrack  & \multirow{2}{*}{106k}&  \multirow{2}{*}{88} &  \multirow{2}{*}{877k}&  \multirow{2}{*}{990}\\
\cite{Sun_2022_CVPR}&&&&\\\hline

 MOT17& \multirow{2}{*}{11k}& \multirow{2}{*}{23} & \multirow{2}{*}{900k}& \multirow{2}{*}{1342}\\
\cite{MOT16}&&&&\\\hline

 MOT20& \multirow{2}{*}{13k}& \multirow{2}{*}{9} & \multirow{2}{*}{2M}& \multirow{2}{*}{3456}\\
\cite{MOTChallenge20}&&&&\\\hline

  UFPR-Planalto801 &\multirow{2}{*}{518k}&\multirow{2}{*}{359}&\multirow{2}{*}{514k}&\multirow{2}{*}{736}\\
 (ours)&&&&\\\hline

\hline
\end{tabular}

\label{tab:mot_datasets}
\end{table}

The \dataset\ dataset was annotated according to the MOTChallenge format and relying on a semi-automatic strategy. People were automatic detected using YOLOv8~\cite{Jocher_YOLO_by_Ultralytics_2023}, which also provided tracking information but with many association errors for our scenario. Thus, we manually reviewed the ID association for every bounding box, in order to assure that one single person is bound to only one ID. This ID consistence is maintained only through one video and a person may appear in another video tied to a different ID. During this review process, bounding boxes showing no people were discarded, and others were corrected as needed.

\section{\uppercase{Proposal}}
\label{sec:proposal}

\noindent
Experiments with state-of-the-art trackers show that one of the main issues with the \dataset\ dataset is the high number of wrongly associated IDs. 
These ID association errors normally occurs when individuals are using similar clothes, performing irregular trajectories, crossing paths and so on. Another cause is high level of tracks fragmentation due to the fact that many people leaves the scene and reappears after a while at an unexpected position.

Considering that, we applied a multilevel strategy, named as \textbf{\supertracker}, where trackers are organized to process different levels of detections. Its general idea is to combine the results of two trackers, referred as trackers Level 1 (L1) and Level 2 (L2). While L1 process every detection, L2 only operates over the ones with high confidence detection scores.   

The selection of the best detections is inspired by the \trade\ Re-ID~\cite{Machaca2022} approach. \trade\ aims to match query images of a person to the corresponding tracklets obtained from video galleries.
After a tracking step for identifying the individual tracklets in the gallery, \trade\ uses an anomaly detection model to find the best representative bounding box of each tracklet, which are then compared to the query image. The top-ranked bounding box is considered the re-identified person or object.

As in \trade, \supertracker\ selects the best bounding box for a tracked ID.
Nevertheless, instead of using a specific model, it uses the detection confidence as an indicative for high-quality detections, since situations prone to ID switch, like occlusions, often produce detections with low confidence scores.

In the \supertracker\ approach, the first tracker (L1) produces the tracklets as in a normal tracking-by-detection operation, by processing the source frame by frame and generating the IDs for detected people in each frame. The result of L1 is a set of detections that compose the L1 tracklets. Each detection is bound to the frame number, box coordinates, confidence score and the associated ID. All these values are buffered to be further processed by the second tracker (L2), which runs in periods of $k$ frames.

After a window of $k$ frames, and considering only the detections in such window,
\supertracker\ selects the bounding box with the highest confidence for each ID.
L2 takes these bounding boxes as input and produces a set with the same format as L1, but with only the top-ranked detection per ID. 

\begin{figure}[ht]
  \centering
   \includegraphics[width=\linewidth]{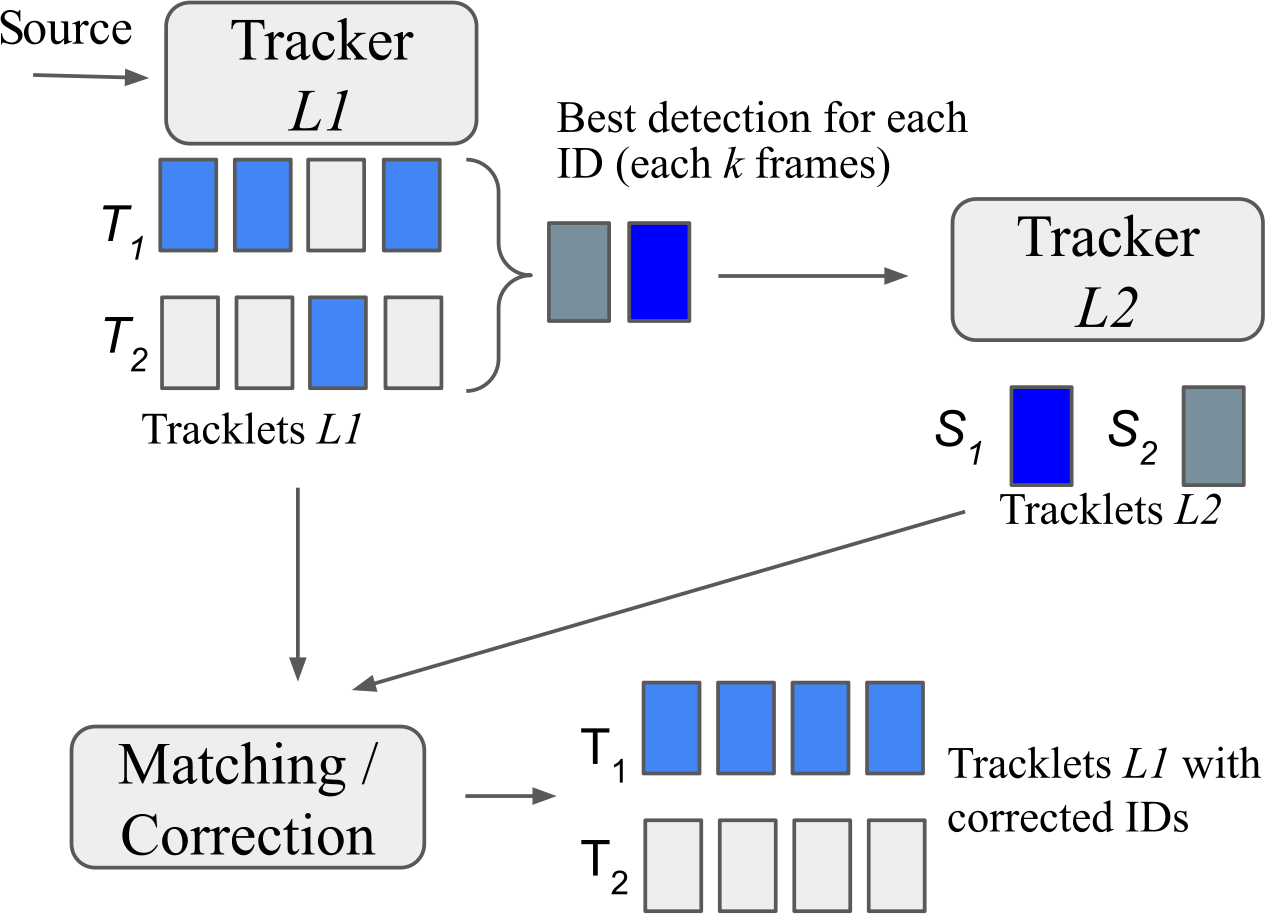}

   \caption{\supertracker\ (proposal) approach overview.}
   \label{fig:proposal}
\end{figure}

\begin{figure}[ht]
  \centering
   \includegraphics[width=.9\linewidth]{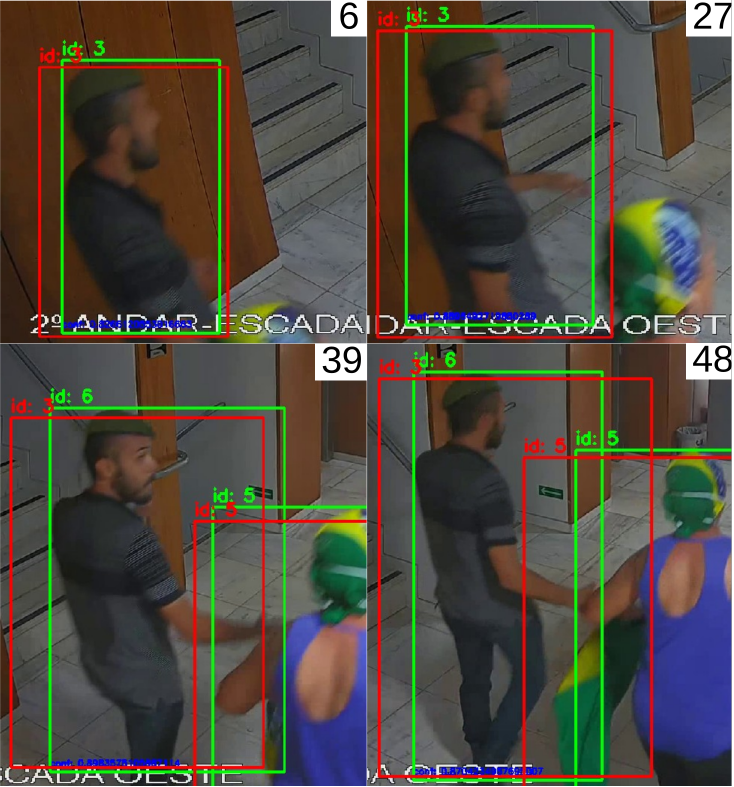}

   \caption{Example of ID correction using \supertracker.}
   \label{fig:id_correction}
\end{figure}

This second level of processing has the effect of creating an alternative tracking state that comprises only high-quality detections, and then it is expected to be less prone to identification errors. 
As an analogy, a wrong ID can be interpreted as noise in the tracklet, so this multilevel approach acts filtering this noise by reducing the sample frequency.
Besides that, this approach also reduces the impact of problems like occlusion and missed detections. As an example, if L1 and L2 are two different instances of the same tracker, L1 holds the tracking state for $n$ frames without a new detection, while L2 will hold it for $k \times n$ frames.

Since L1 and L2 are two distinct instances, there is no correspondence between the IDs they provide. This may result in a single person being associated to one ID by L1 and to a different ID by L2.
Therefore, \supertracker\ implements an additional step for matching the IDs taking L2 as reference.
Such step also acts to correct the remaining L1 detections by replacing them by the matches in L2.

For matching, \supertracker\ computes a matrix of IoU distances per frame in the window. This matrix relates the L1 detections in the frame to L2 detections.
The matching is solved by using the Jonker-Volgenant algorithm~\cite{jonker1987} for linear assignment.
Notice that in L2 there is only one detection for each ID, therefore these detections are re-used when computing the matrix for all the frames in the window. 
\Cref{fig:proposal} presents an overview of the proposed approach. In this figure, $T_1$ and $T_2$ represent the L1 tracklets, while $S_1$ and $S_2$ are the L2 detections.

\Cref{fig:id_correction} presents an example of ID correction using \supertracker. L1 detections are in green, while L2 detections are in red. In frames $6$ and $27$ both L1 and L2 associate a person to the ID $3$.  In frame $27$ L1 wrongly associates the ID $6$ to the same person, keeping this new ID in subsequent frames. L2, however, is more consistent and keeps the same ID through this video sequence.
Notice that there is no apparent situation to justify this wrong association by L1.

\section{\uppercase{Experiments Setup}}
\label{sec:setup}

\noindent
Despite \supertracker\ being agnostic in terms of what trackers can be used as L1 or L2, for the context of this paper we evaluated this approach using six possible trackers as L1 and two as L2. Each of these were combined to compose the $12$ different pairs of trackers used in the experiments.
Besides that, the experiments also considered different values for $k$ in order to assess the impact of the frame window size for running the tracker L2. We considered $k \in \{2,3,5,10\}$.

For L1 we selected \bytetrack, \botsort, \strongsort, \ocsort, \deepocsort\ and \hybridsort.
For L2 we selected \bytetrack\ and \ocsort. Since neither \bytetrack\ nor \ocsort\ require re-identification models to compute features, their use as L2 helps to reduce the overhead required for buffering detections, since there is no need to keep frames or appearance features while waiting for L2 to run. 
The source code for these trackers is provided by BoxMOT~\cite{Brostrom_BoxMOT_A_collection}.
For trackers that require a Re-ID model, we used pre-trained models for OSNet~\cite{Zhou2019} and CLIP-ReID~\cite{Li_Sun_Li_2023}, also provided by BoxMOT.%

\supertracker\ uses YOLOv8~\cite{Jocher_YOLO_by_Ultralytics_2023} for people detection. This version is a variant of YOLO~\cite{Redmon_2016_CVPR}, a model widely used in computer vision tasks.
YOLOv8 provides pre-trained models, including the versions \textit{nano}, \textit{small}, \textit{medium}, \textit{large}, and \textit{extra-large}. 
Notice that the \textit{small} version of YOLOv8 is the one used in \supertracker\, despite it is not the best in terms of precision in comparison to larger YOLOv8 models. However, it is the second fastest while still presenting good qualitative results, what makes it acceptable in our context.


\subsection{Evaluation Metrics}

\noindent
The metrics considered for evaluation of \supertracker\ are the IDF1 Score~\cite{Ristani2016}, Higher Order Tracking Accuracy (HOTA)~\cite{luiten2020IJCV}, Multi-Object Tracking Accuracy~(MOTA) and Multi-Object Tracking Precision~(MOTP)~\cite{clear2008}. All these metrics are widely used in the evaluation of object tracking algorithms and were computed by using TrackEval~\cite{luiten2020trackeval}.

\begin{equation}
\label{eq:mota}
    MOTA = 1 - \frac{|FN| + |FP| + |IDSW|}{|gtDet|} 
\end{equation}

\begin{equation}
\label{eq:motp}
    MOTP = \frac{1}{|TP|} \sum_{TP}S 
\end{equation}

\begin{equation}
\label{eq:idf1}
    IDF1 = \frac{|IDTP|}{|IDTP| + 0.5|IDFN| + 0.5|IDFP|}  
\end{equation}

\begin{equation}
\label{eq:hota}
    HOTA = \int_{0}^{1} \sqrt{DetA_{\alpha}.AssA_{\alpha}} \,dx
\end{equation}

$MOTA$ relates the number of ground-truth detections ($gtDet$) and the numbers of false positives ($FP$), false negatives ($FN$), and ID switches ($IDSW$), as showed in \cref{eq:mota}.
$MOTP$ computes the overall mean error between the estimated and detected positions, as in \cref{eq:motp}, where $S$ is a measure of similarity.

$IDF1$ is used to assess the accuracy of maintaining consistent object identities over predicted trajectories related to the ground truth. $IDF1$ is computed according to \cref{eq:idf1}, where $IDTP$ (Identity True Positives) is the number of correctly associated identities, $IDFN$ (Identity False Negatives) is the number of missed IDs and $IDFP$ (Identity False Positives) is the number of IDs in trajectories not present in the ground truth.

$HOTA$ is a comprehensive metric encompassing a set of other metrics derived from $MOTA$ and $MOTP$. It is designed to assess the accuracy of localization, detection, and association.  As shown in \cref{eq:hota}, it involves integrating geometric means for detection ($DetA$) and association ($AssA$) scores, calculated for a range of thresholds ($\alpha$, where $0 \leq \alpha \leq 1$).

$HOTA$, $MOTA$ and $MOTP$ give us an general overview of the proposal efficiency. However, since our approach is focused on the ID association problem, we consider $IDF1$ more relevant to this context.

\section{\uppercase{Results}}
\label{sec:results}

\noindent
This section presents the results from combining twelve pairs of trackers L1/L2. 
Each tracker was also evaluated without using the \supertracker\ scheme in order to establish the baselines for comparison. 
For a matter of clarity, we will address to L1 as the "base tracker" when referring to it running solo.  
The results are presented considering a pair in comparison to its respective base tracker.
Tables \ref{tab:bytetrack} to \ref{tab:deepocsort} show the scored values for all combinations. Tables regarding trackers that use a Re-ID model include the results for the models we used, \osnet\ and \clip.
Notice that there is no info regarding L2 or $k$ in the top data row of each of those tables, since these rows show the results for the base tracker. All score values are presented as percentage, and the higher values are better. 

Overall, the cases with $k=2$ or $k=3$ are the ones with the higher \idfone, \hota\ and \mota\ scores, while cases with $k=10$ perform best considering \motp. However, \motp\ typically has a very little variation (from $0.1\%$ to $0.4\%$) within the set of combinations for the same base tracker, resulting in no real benefit when using $k=10$.
\bytetrack\ as L2 generally performs better than \ocsort, which shows the worst results when paired to \strongsort\ or \hybridsort. In such cases, the base tracker outperforms any combination in terms of \idfone\ and \hota\ scores, regardless the value of $k$ or the re-id model. The exceptions are the cases where \hybridsort\ is using \clip\ and even so there is only a maximum gain of $0.2\%$ for pairing \ocsort\ to \hybridsort.

\bytetrack\ is the tracker among all base trackers who performs better. Despite of that, \supertracker\ was able to achieve higher values in every metric for several combinations.
When using \bytetrack\ as L1, the best \hota\ score occurs when using \ocsort\ as L2 and $k=2$. This case achieved a \hota\ score of $41.7\%$. This is also the highest \hota\ score in comparison to all other trackers combinations.
In terms of \idfone, \supertracker\ shows an improvement from $41.1\%$ to $43.2\%$ regarding the base tracker.
\Cref{tab:bytetrack} presents the results for \bytetrack\ used as L1.

Opposite to \bytetrack, \ocsort\ is the base tracker with the lowest values. As shown in \Cref{tab:ocsort},  its base values for \mota, \idfone\ and \hota\ are $51.2\%$, $33.0\%$ and $35.4\%$, respectively.
Its performance improved when applying the \supertracker\ strategy. The use of \bytetrack\ as L2 resulted in values up to $65.1\%$, $42.3\%$ and $41.0\%$ for \mota, \idfone\ and \hota, respectively.  
These new values constitute the larger improvement achieved for \hota\ score among all assessed base trackers.
Using \ocsort\ as both L1 and L2 granted lower scores, but it still provides some improvement in cases where $k \in \{2,3\}$.

\Cref{tab:botsort} shows that \botsort\ achieved balanced numbers regarding the use of \bytetrack\ or \ocsort\ as L2, with no accentuated 
difference between the cases where $k \in \{2,3,5\}$.
Despite of that, it also achieved best values with \supertracker, except in cases where $k=10$.
Best \hota\ and \motp\ values for \botsort\ are $38.0\%$ and $81.6\%$, respectively, which occurred for both re-identification models. \mota\ score was higher with \osnet\ ($46.3\%$) while \clip\ achieved $39.3\%$ for \idfone\, overcoming the base tracker score of $37.0\%$.

\begin{table}[ht]
\fontsize{9pt}{9pt}\selectfont
    \centering
\addtolength{\tabcolsep}{-4.5pt}

\caption{Results for \bytetrack\ as L1.}

\begin{tabular}{|l|C|CCCC|}
\hline
\multicolumn{2}{|l|}{Tracker L1} & \multicolumn{4}{l|}{\bytetrack}    \\
\hline
 Tracker L2 & k & IDF1 & HOTA & MOTA & MOTP \\
\hline
 $-$ & - & 41.1 & 41.2 & 64.6 & 82.7 \\
 \bytetrack & 2 & 42.5 & 41.4 & 65.0 & 82.8 \\
 \bytetrack & 3 & \textbf{43.2} & 41.6 & 65.2 & 82.8 \\
 \bytetrack & 5 & 42.4 & 40.6 & \textbf{65.3} & 82.9 \\
 \bytetrack & 10 & 35.9 & 35.1 & 63.5 & \textbf{83.0} \\
 \ocsort & 2 & 42.8 & \textbf{41.7} & 64.8 & 82.8 \\
 \ocsort & 3 & 42.9 & 41.6 & 64.9 & 82.8 \\
 \ocsort & 5 & 42.6 & 40.8 & 64.8 & 82.8 \\
 \ocsort & 10 & 34.8 & 34.2 & 63.4 & 82.9 \\
\hline
\end{tabular}
\label{tab:bytetrack}
\end{table}

\begin{table}[ht!]
\fontsize{9pt}{9pt}\selectfont
\centering
\addtolength{\tabcolsep}{-4.5pt}
\caption{Results for \ocsort\ as L1.}

\begin{tabular}{|l|C|CCCC|}
\hline
\multicolumn{2}{|l|}{Tracker L1} & \multicolumn{4}{l|}{\ocsort}    \\
\hline
 Tracker L2 & k & IDF1 & HOTA & MOTA & MOTP \\
\hline
 $-$ & - & 33.0 & 35.4 & 51.2 & 81.3 \\
 \bytetrack & 2 & \textbf{42.3} & \textbf{41.0} & \textbf{65.1} & 81.6 \\
 \bytetrack & 3 & 41.9 & 40.5 & 65.0 & 81.6 \\
 \bytetrack & 5 & 41.3 & 39.9 & 64.9 & 81.6 \\
 \bytetrack & 10 & 34.0 & 33.6 & 62.8 & \textbf{81.7} \\
 \ocsort & 2 & 34.3 & 35.8 & 52.1 & 81.3 \\
 \ocsort & 3 & 34.1 & 35.5 & 52.1 & 81.3 \\
 \ocsort & 5 & 33.4 & 34.3 & 51.9 & 81.3 \\
 \ocsort & 10 & 28.3 & 29.6 & 50.6 & 81.4 \\
\hline
\end{tabular}
\label{tab:ocsort}
\end{table}

\begin{table}[ht]
\fontsize{9pt}{9pt}\selectfont
    \centering
\addtolength{\tabcolsep}{-4.5pt}
\caption{Results for \botsort\ as L1.}
\begin{tabular}{|l|C|C|C|C|C|C|C|C|C|}
\hline
\multicolumn{2}{|l|}{Tracker L1} & \multicolumn{8}{c|}{\botsort}    \\
\hline
\multicolumn{2}{|l|}{Re-ID} & \multicolumn{4}{c|}{OSNet} & \multicolumn{4}{c|}{CLIP-ReID}   \\
\hline
Tracker L2 & $k$ & \rotatebox[origin=c]{90}{IDF1} & \rotatebox[origin=c]{90}{HOTA} & \rotatebox[origin=c]{90}{MOTA} & \rotatebox[origin=c]{90}{MOTP} & \rotatebox[origin=c]{90}{IDF1} & \rotatebox[origin=c]{90}{HOTA} & \rotatebox[origin=c]{90}{MOTA} & \rotatebox[origin=c]{90}{MOTP}\\

\hline
 $-$ &  - & 36.8 & 37.1 & 42.0 & 81.3 & 37.0 & 37.4 & 41.2 & 81.3 \\
\bytetrack & 2 & 38.2 & 37.5 & 46.0 & 81.5 & 38.6 & 37.9 & 45.3 & 81.5 \\
\bytetrack & 3 & \textbf{38.9} & 37.7 & 46.1 & 81.4 & \textbf{39.3} & \textbf{38.0} & 45.4 & 81.4 \\
\bytetrack & 5 & 38.2 & 37.2 & \textbf{46.3} & 81.5 & 38.3 & 37.2 & \textbf{45.5} & 81.5 \\

\bytetrack & 10 & 33.0 & 32.2 & 45.4 & \textbf{81.6} & 32.6 & 31.9 & 44.6 & \textbf{81.6} \\
\ocsort & 2 & 38.5 & \textbf{38.0} & 42.5 & 81.3 & 38.1 & 37.9 & 41.8 & 81.3 \\
\ocsort & 3 & 38.1 & 37.2 & 42.6 & 81.3 & 37.7 & 37.0 & 41.9 & 81.3 \\
\ocsort & 5 & 37.6 & 36.4 & 42.7 & 81.3 & 37.7 & 36.5 & 41.9 & 81.3 \\
\ocsort & 10 & 31.3 & 31.0 & 41.9 & 81.4 & 30.4 & 30.7 & 41.1 & 81.4 \\

\hline
\end{tabular}

\label{tab:botsort}
\end{table}

\begin{table}[ht]
\fontsize{9pt}{9pt}\selectfont
\addtolength{\tabcolsep}{-4.5pt}

    \centering
\caption{Results for \strongsort\ as L1.}
\begin{tabular}{|l|C|C|C|C|C|C|C|C|C|}
\hline
\multicolumn{2}{|l|}{Tracker L1} & \multicolumn{8}{c|}{\strongsort}    \\
\hline
\multicolumn{2}{|l|}{Re-ID} & \multicolumn{4}{c|}{OSNet} & \multicolumn{4}{c|}{CLIP-ReID}   \\
\hline
Tracker L2 & $k$ & \rotatebox[origin=c]{90}{IDF1} & \rotatebox[origin=c]{90}{HOTA} & \rotatebox[origin=c]{90}{MOTA} & \rotatebox[origin=c]{90}{MOTP} & \rotatebox[origin=c]{90}{IDF1} & \rotatebox[origin=c]{90}{HOTA} & \rotatebox[origin=c]{90}{MOTA} & \rotatebox[origin=c]{90}{MOTP}\\

\hline

 $-$ & - & 35.9 & 37.3 & 52.2 & 81.3 & 35.8 & 37.4 & 52.3 & 81.4 \\
 \bytetrack & 2 & 42.4 & 40.8 & \textbf{65.3} & 81.6 & 42.5 & 41.0 & \textbf{65.3} & \textbf{81.7} \\
 \bytetrack & 3 & \textbf{42.9} & \textbf{41.1} & 65.2 & 81.6 & \textbf{42.6} & \textbf{41.1} & 65.2 & 81.6 \\
 \bytetrack & 5 & 41.2 & 39.9 & 65.1 & 81.6 & 41.9 & 40.5 & 65.1 & \textbf{81.7} \\
 \bytetrack & 10 & 34.9 & 34.2 & 63.0 & \textbf{81.7} & 34.7 & 33.8 & 63.0 & \textbf{81.7} \\

 \ocsort & 2 & 35.4 & 36.5 & 53.1 & 81.4 & 35.4 & 36.5 & 53.2 & 81.4 \\
 \ocsort & 3 & 35.9 & 36.8 & 53.1 & 81.4 & 35.4 & 36.3 & 53.3 & 81.4 \\
 \ocsort & 5 & 34.5 & 35.0 & 52.9 & 81.4 & 34.1 & 34.8 & 53.0 & 81.4 \\
 \ocsort & 10 & 28.4 & 29.7 & 51.5 & 81.4 & 28.5 & 29.7 & 51.6 & 81.5 \\
 
\hline
\end{tabular}
\label{tab:strongsort}
\end{table}

\begin{table}[ht]
\fontsize{9pt}{9pt}\selectfont
\addtolength{\tabcolsep}{-4.5pt}

    \centering
\caption{Results for \hybridsort\ as L1.}
\begin{tabular}{|l|C|C|C|C|C|C|C|C|C|}
\hline
\multicolumn{2}{|l|}{Tracker L1} & \multicolumn{8}{c|}{\hybridsort}    \\
\hline
\multicolumn{2}{|l|}{Re-ID} & \multicolumn{4}{c|}{OSNet} & \multicolumn{4}{c|}{CLIP-ReID}   \\
\hline
Tracker L2 & $k$ & \rotatebox[origin=c]{90}{IDF1} & \rotatebox[origin=c]{90}{HOTA} & \rotatebox[origin=c]{90}{MOTA} & \rotatebox[origin=c]{90}{MOTP} & \rotatebox[origin=c]{90}{IDF1} & \rotatebox[origin=c]{90}{HOTA} & \rotatebox[origin=c]{90}{MOTA} & \rotatebox[origin=c]{90}{MOTP}\\

\hline

 $-$ & - & 34.6 & 36.0 & 50.9 & 81.3 & 34.1 & 36.0 & 51.2 & 81.3 \\
 \bytetrack & 2 & \textbf{42.5} & \textbf{41.0} & \textbf{65.1} & 81.6 & 42.5 & 41.0 & \textbf{65.1} & 81.6 \\
 \bytetrack & 3 & 42.2 & 40.6 & \textbf{65.1} & 81.6 & \textbf{43.1} & \textbf{41.2} & 65.0 & 81.6 \\
 \bytetrack & 5 & 41.9 & 40.0 & 65.0 & 81.6 & 41.9 & 40.1 & 65.0 & 81.6 \\
  \bytetrack & 10 & 34.0 & 33.6 & 63.0 & \textbf{81.7} & 34.6 & 33.8 & 62.9 & \textbf{81.7} \\

 \ocsort & 2 & 33.9 & 35.8 & 51.8 & 81.3 & 34.3 & 36.0 & 52.0 & 81.3 \\
 \ocsort & 3 & 34.2 & 35.6 & 51.9 & 81.3 & 34.3 & 35.7 & 52.1 & 81.3 \\
 \ocsort & 5 & 34.1 & 34.8 & 51.8 & 81.3 & 34.2 & 34.9 & 51.9 & 81.3 \\
 \ocsort & 10 & 28.4 & 29.5 & 50.6 & 81.4 & 27.9 & 29.3 & 50.6 & 81.4 \\

\hline
\end{tabular}

\label{tab:hybridsort}
\end{table}

\begin{table}[ht]
\fontsize{9pt}{9pt}\selectfont
\addtolength{\tabcolsep}{-4.5pt}

    \centering
\caption{Results for \deepocsort\ as L1.}
\begin{tabular}{|l|C|C|C|C|C|C|C|C|C|}
\hline
\multicolumn{2}{|l|}{Tracker L1} & \multicolumn{8}{c|}{\deepocsort}    \\
\hline
\multicolumn{2}{|l|}{Re-ID} & \multicolumn{4}{c|}{OSNet} & \multicolumn{4}{c|}{CLIP-ReID}   \\
\hline
Tracker L2 & $k$ & \rotatebox[origin=c]{90}{IDF1} & \rotatebox[origin=c]{90}{HOTA} & \rotatebox[origin=c]{90}{MOTA} & \rotatebox[origin=c]{90}{MOTP} & \rotatebox[origin=c]{90}{IDF1} & \rotatebox[origin=c]{90}{HOTA} & \rotatebox[origin=c]{90}{MOTA} & \rotatebox[origin=c]{90}{MOTP}\\

\hline

 $-$ & - & 35.0 & 36.7 & 62.1 & 81.7 & 35.6 & 37.3 & 62.2 & 81.7 \\
 \bytetrack & 2 & 43.5 & 41.4 & 65.5 & 81.9 & 43.6 & 41.4 & 65.5 & 81.9 \\
 \bytetrack & 3 & \textbf{44.5} & \textbf{41.7} & \textbf{65.6 }& 81.9 & \textbf{44.6} & \textbf{41.7} & \textbf{65.6} & 81.9 \\
 \bytetrack & 5 & 42.6 & 40.3 & 65.4 & 81.9 & 42.9 & 40.5 & 65.5 & 81.9 \\
  \bytetrack & 10 & 35.0 & 33.9 & 63.6 & \textbf{82.0} & 35.1 & 33.8 & 63.6 & \textbf{82.0} \\

 \ocsort & 2 & 41.0 & 39.7 & 62.7 & 81.7 & 41.3 & 39.9 & 62.7 & 81.7 \\
 \ocsort & 3 & 39.9 & 39.4 & 62.7 & 81.7 & 40.1 & 39.5 & 62.7 & 81.7 \\
 \ocsort & 5 & 38.3 & 37.4 & 62.4 & 81.8 & 38.5 & 37.6 & 62.4 & 81.8 \\
  \ocsort & 10 & 31.1 & 31.4 & 60.9 & 81.8 & 31.7 & 31.8 & 61.0 & 81.8 \\

\hline
\end{tabular}
\label{tab:deepocsort}
\end{table}

The results for \strongsort\ with \osnet\ and \clip\ are very close, as \Cref{tab:strongsort} shows. This occurs in all combinations, despite using or not the \supertracker\ strategy. Best values for \idfone\ and \hota\ are with $k=3$, achieving, respectively, $42.9\%$ and $41.1\%$. Best value for \mota\ overcomes the base tracker in more than $13\%$.
These cases used \bytetrack\ as L2.
When using \ocsort\ as L2, however, \strongsort\ improved only for \mota\ and in fact performed slightly worst than the base tracker. By observing the results we noticed almost $2 \times$ more ID switches when \strongsort\ was paired to \ocsort\ than when paired to \bytetrack. It is important to point that this high number of ID switches is close to what occurs for \strongsort\ with no pairing.

Similar to \strongsort, \hybridsort\ performed a high number of ID switches and did not achieve good results with \ocsort\ as L2. 
\hybridsort\ inherits ideas from both \strongsort\ and \ocsort, which do not present improvements when using \ocsort\ as L2. The reasons for that are still unclear, but, as a hypothesis, we believe that since they all originally already try to deal with long-term associations, relying on virtual trajectories with sparse detection from \ocsort\ have none to little effect.
When using \bytetrack\ as L2 there is a gain in every metric for $k \in \{2,3,5\}$, as one can see in \Cref{tab:hybridsort}.
Best \idfone\ score achieved is $43.1\%$ while for \hota\ and \mota, respectively the best values are $41.2\%$ and $65.1\%$.

\deepocsort\ achieved very close results with both \osnet\ and \clip\ (please refer to \Cref{tab:deepocsort}). It ties to \bytetrack\ in terms of \hota ($41.7\%$), but it is better in terms of \idfone\ ($44.5\%$) and \mota\ ($65.6$).
\deepocsort\ is the base tracker that most benefits from the \supertracker\ approach. It reached a maximum gain of $9.5\%$ in \idfone. In terms of L2, \deepocsort\ pairs show higher scores in comparison to the base tracker, for both \bytetrack\ and \ocsort.

\section{Conclusion}
\label{sec:conclusion}
\noindent
This paper presented the \supertracker, a pairing strategy that aims to improve people tracking by correcting ID association. \supertracker\ was applied to the \dataset\ dataset, which was created from public videos on invasions to the Palácio do Planalto, in Brazil, occurred at January 8, 2023.
The presented results show that the use of this pairing technique provides better results in comparison to using only individual trackers.

As future work, we intend to extend the dataset and to provide annotations for it to be suitable to other Computer Vision tasks, such as gait recognition and action detection. 
We also aim to improve the proposed strategy by enabling the use of level 2 trackers that support deep learning feature extraction models.

\section*{ACKNOWLEDGEMENTS}

This work was supported in part by the Coordination for the Improvement of Higher Education Personnel~(CAPES) (\textit{Programa de Cooperação Acad\^{e}mica em Seguran\c{c}a P\'{u}blica e Ci\^{e}ncias Forenses \#~88881.516265/2020-01}), and in part by the National Council for Scientific and Technological Development(CNPq) (\#~308879/2020-1).
We gratefully acknowledge the support of NVIDIA Corporation with the donation of the Quadro RTX $8000$ GPU used for this research.

\bibliographystyle{apalike}
{\small
\bibliography{references}}

\begin{thebibliography}{}

\bibitem[Aharon et~al., 2022]{aharon2022bot}
Aharon, N., Orfaig, R., and Bobrovsky, B.-Z. (2022).
\newblock {BoT-SORT: Robust Associations Multi-Pedestrian Tracking}.
\newblock {\em arXiv:2206.14651}.

\bibitem[Bernardin and Stiefelhagen, 2008]{clear2008}
Bernardin, K. and Stiefelhagen, R. (2008).
\newblock {Evaluating Multiple Object Tracking Performance: The CLEAR MOT Metrics}.
\newblock {\em J. Image Video Process.}, 2008.

\bibitem[Bewley et~al., 2016]{Bewley2016}
Bewley, A., Ge, Z., Ott, L., Ramos, F., and Upcroft, B. (2016).
\newblock Simple online and realtime tracking.
\newblock In {\em 2016 IEEE Int. Conf. on Image Processing (ICIP)}, pages 3464--3468.

\bibitem[Broström, 2023]{Brostrom_BoxMOT_A_collection}
Broström, M. (2023).
\newblock {BoxMOT: A collection of SOTA real-time, multi-object trackers for object detectors }.

\bibitem[Cao et~al., 2023]{Cao_2023_CVPR}
Cao, J., Pang, J., Weng, X., Khirodkar, R., and Kitani, K. (2023).
\newblock {Observation-Centric SORT: Rethinking SORT for Robust Multi-Object Tracking}.
\newblock In {\em Proceedings of the IEEE/CVF Conference on Computer Vision and Pattern Recognition (CVPR)}, pages 9686--9696.

\bibitem[Dendorfer et~al., 2019]{MOT19_CVPR}
Dendorfer, P., Rezatofighi, H., Milan, A., Shi, J., Cremers, D., Reid, I., Roth, S., Schindler, K., and Leal-Taix\'{e}, L. (2019).
\newblock {CVPR19} tracking and detection challenge: {H}ow crowded can it get?
\newblock {\em arXiv:1906.04567 [cs]}.

\bibitem[Dendorfer et~al., 2020]{MOTChallenge20}
Dendorfer, P., Rezatofighi, H., Milan, A., Shi, J., Cremers, D., Reid, I., Roth, S., Schindler, K., and Leal-Taix\'{e}, L. (2020).
\newblock {MOT20: A benchmark for multi object tracking in crowded scenes}.
\newblock {\em arXiv:2003.09003[cs]}.

\bibitem[Du et~al., 2021]{GIAOTracker2021}
Du, Y., Wan, J., Zhao, Y., Zhang, B., Tong, Z., and Dong, J. (2021).
\newblock {GIAOTracker: A comprehensive framework for MCMOT with global information and optimizing strategies in VisDrone 2021}.
\newblock In {\em 2021 IEEE/CVF Int. Conf. on Computer Vision Workshops (ICCVW)}, pages 2809--2819.

\bibitem[Du et~al., 2023]{du2023strongsort}
Du, Y., Zhao, Z., Song, Y., Zhao, Y., Su, F., Gong, T., and Meng, H. (2023).
\newblock {StrongSORT: Make DeepSORT Great Again}.
\newblock {\em IEEE Transactions on Multimedia}.

\bibitem[{Gabinete de Segurança Institucional}, 2023]{GSI2023}
{Gabinete de Segurança Institucional} (2023).
\newblock Nota à imprensa.
\newblock Last accessed 25 September 2023.

\bibitem[Ge et~al., 2021]{ge2021yolox}
Ge, Z., Liu, S., Wang, F., Li, Z., and Sun, J. (2021).
\newblock {YOLOX: Exceeding YOLO Series in 2021}.

\bibitem[Jocher et~al., 2023]{Jocher_YOLO_by_Ultralytics_2023}
Jocher, G., Chaurasia, A., and Qiu, J. (2023).
\newblock {YOLO by Ultralytics}.

\bibitem[Jonker and Volgenant, 1987]{jonker1987}
Jonker, R. and Volgenant, A. (1987).
\newblock {A Shortest Augmenting Path Algorithm for Dense and Sparse Linear Assignment Problems}.
\newblock {\em {Computing}}, 38(4):325–340.

\bibitem[Li et~al., 2023]{Li_Sun_Li_2023}
Li, S., Sun, L., and Li, Q. (2023).
\newblock {CLIP-ReID: Exploiting Vision-Language Model for Image Re-identification without Concrete Text Labels}.
\newblock {\em Proc. of the AAAI Conf. on Artificial Intelligence}, 37(1):1405--1413.

\bibitem[Luiten and Hoffhues, 2020]{luiten2020trackeval}
Luiten, J. and Hoffhues, A. (2020).
\newblock {TrackEval}.
\newblock {https://github.com/JonathonLuiten/TrackEval}.

\bibitem[Luiten et~al., 2020]{luiten2020IJCV}
Luiten, J., Osep, A., Dendorfer, P., Torr, P., Geiger, A., Leal-Taix{\'e}, L., and Leibe, B. (2020).
\newblock {HOTA: A Higher Order Metric for Evaluating Multi-Object Tracking}.
\newblock {\em Int. Journal of Computer Vision}, pages 1--31.

\bibitem[Machaca et~al., 2022]{Machaca2022}
Machaca, L., Sumari~H., F.~O., Huaman, J., Clua, E., and Guerin, J. (2022).
\newblock {TrADe Re-ID – Live Person Re-Identification using Tracking and Anomaly Detection}.
\newblock In {\em 2022 21st IEEE Int. Conf. on Machine Learning and Applications (ICMLA)}, pages 449--454.

\bibitem[Maggiolino et~al., 2023]{maggiolino2023deep}
Maggiolino, G., Ahmad, A., Cao, J., and Kitani, K. (2023).
\newblock {Deep OC-SORT: Multi-Pedestrian Tracking by Adaptive Re-Identification}.
\newblock {\em arXiv:2302.11813}.

\bibitem[Milan et~al., 2016]{MOT16}
Milan, A., Leal-Taix\'{e}, L., Reid, I., Roth, S., and Schindler, K. (2016).
\newblock {MOT}16: {A} benchmark for multi-object tracking.
\newblock {\em arXiv:1603.00831 [cs]}.

\bibitem[Redmon et~al., 2016]{Redmon_2016_CVPR}
Redmon, J., Divvala, S., Girshick, R., and Farhadi, A. (2016).
\newblock You only look once: Unified, real-time object detection.
\newblock In {\em Proceedings of the IEEE Conference on Computer Vision and Pattern Recognition (CVPR)}.

\bibitem[Ristani et~al., 2016]{Ristani2016}
Ristani, E., Solera, F., Zou, R., Cucchiara, R., and Tomasi, C. (2016).
\newblock {Performance Measures and a Data Set for Multi-target, Multi-camera Tracking}.
\newblock In {\em Computer Vision -- ECCV 2016 Workshops}, pages 17--35, Cham. Springer International Publishing.

\bibitem[Sun et~al., 2022]{Sun_2022_CVPR}
Sun, P., Cao, J., Jiang, Y., Yuan, Z., Bai, S., Kitani, K., and Luo, P. (2022).
\newblock {DanceTrack: Multi-Object Tracking in Uniform Appearance and Diverse Motion}.
\newblock In {\em Proceedings of the IEEE/CVF Conference on Computer Vision and Pattern Recognition (CVPR)}, pages 20993--21002.

\bibitem[Wojke et~al., 2017]{DeepSORT2017}
Wojke, N., Bewley, A., and Paulus, D. (2017).
\newblock Simple online and realtime tracking with a deep association metric.
\newblock In {\em 2017 IEEE Int. Conf. on Image Processing (ICIP)}, pages 3645--3649.

\bibitem[Yang et~al., 2023]{yang2023hybrid}
Yang, M., Han, G., Yan, B., Zhang, W., Qi, J., Lu, H., and Wang, D. (2023).
\newblock {Hybrid-SORT: Weak Cues Matter for Online Multi-Object Tracking}.
\newblock {\em arXiv:2308.00783}.

\bibitem[Zhang et~al., 2022]{zhang2022bytetrack}
Zhang, Y., Sun, P., Jiang, Y., Yu, D., Weng, F., Yuan, Z., Luo, P., Liu, W., and Wang, X. (2022).
\newblock {ByteTrack: Multi-Object Tracking by Associating Every Detection Box}.
\newblock In {\em Proceedings of the European Conference on Computer Vision (ECCV)}.

\bibitem[Zhou et~al., 2019]{Zhou2019}
Zhou, K., Yang, Y., Cavallaro, A., and Xiang, T. (2019).
\newblock {Omni-Scale Feature Learning for Person Re-Identification}.
\newblock In {\em 2019 IEEE/CVF Int. Conf. on Computer Vision (ICCV)}, pages 3701--3711, Los Alamitos, CA, USA. IEEE Computer Society.

\end{thebibliography}

\end{document}